\documentclass[letterpaper]{article} 
\usepackage{aaai2026}  
\usepackage{times}  
\usepackage{helvet}  
\usepackage{courier}  
\usepackage[hyphens]{url}  
\usepackage{graphicx} 
\usepackage{natbib}  
\usepackage{caption} 
\usepackage{amsfonts,amssymb}
\usepackage{amsmath} 

\usepackage{algorithm}
\usepackage{algorithmic}
\usepackage[table]{xcolor}

\usepackage{makecell}
\usepackage{multirow}
\usepackage{pifont}
\usepackage{xcolor}
\usepackage{booktabs}
\usepackage{graphicx}
\newcommand{\cmark}{\textcolor{green!60!black}{\ding{51}}} 
\newcommand{\xmark}{\textcolor{red}{\ding{55}}} 

\usepackage{newfloat}
\usepackage{listings}
\usepackage{newfloat}
\usepackage{listings}
\usepackage{array} 
\usepackage{algorithm}
\usepackage{algorithmic}

\usepackage{newfloat}
\usepackage{listings}
\DeclareCaptionStyle{ruled}{labelfont=normalfont,labelsep=colon,strut=off} 
\floatstyle{ruled}
\newfloat{listing}{tb}{lst}{}
\floatname{listing}{Listing}
\title{Hybrid-DMKG: A Hybrid Reasoning Framework over Dynamic Multimodal Knowledge Graphs for Multimodal Multihop QA with Knowledge Editing}
\author{
Li Yuan\textsuperscript{\rm 1,\rm2}, 
Qingfei Huang\textsuperscript{\rm 1,\rm2}, 
Bingshan Zhu\textsuperscript{\rm 3}, 
Yi Cai\textsuperscript{\rm1,\rm2}\thanks{Corresponding author: Yi Cai (ycai@scut.edu.cn)}, 
Qingbao Huang\textsuperscript{\rm4},
Changmeng Zheng \textsuperscript{\rm5},
Zikun Deng\textsuperscript{\rm 1,\rm2},
Tao Wang\textsuperscript{\rm 6}
}
\affiliations{
    \textsuperscript{\rm 1}School of Software Engineering, South China University of Technology, Guangzhou, China\\
    \textsuperscript{\rm 2} Key Laboratory of Big Data and Intelligent Robot (SCUT), MOE of China\\
    \textsuperscript{\rm 3} School of Big Data and Artificial Intelligence, Guangdong University of Finance \& Economics\\
     \textsuperscript{\rm 4} School of Electrical Engineering, Guangxi University, Nanning, Guangxi, China\\
      \textsuperscript{\rm 5} Department of Computing, The Hong Kong Polytechnic University, Hong Kong, China\\
    
    \textsuperscript{\rm 6} Department of Biostatistics \& Health Informatics, King's College London, London, United Kingdom \\

  \{seyuanli@mail,ycai@,zkdeng@\}.scut.edu.cn, qbhuang@gxu.edu.cn, changmeng.zheng@polyu.edu.hk, tao.wang@kcl.ac.uk

}

\usepackage{bibentry}

\begin{document}

\maketitle

\begin{abstract}
Multimodal Knowledge Editing (MKE) extends traditional knowledge editing to settings involving both textual and visual modalities. However, existing MKE benchmarks primarily assess final answer correctness, neglecting the quality of intermediate reasoning and robustness to visually rephrased inputs. To address this limitation, we introduce MMQAKE, the first benchmark for multimodal multihop question answering with knowledge editing. MMQAKE evaluates: (1) a model’s ability to reason over 2–5-hop factual chains that span both text and images, including performance at each intermediate step; (2) robustness to visually rephrased inputs in multihop questions.
Our evaluation shows that current MKE methods often struggle to consistently update and reason over multimodal reasoning chains following knowledge edits. 
To overcome these challenges, we propose Hybrid-DMKG, a hybrid reasoning framework built on a dynamic multimodal knowledge graph (DMKG) to enable accurate multihop reasoning over updated multimodal knowledge. Hybrid-DMKG first uses a large language model to decompose multimodal multihop questions into sequential sub-questions, then applies a multimodal retrieval model to locate updated facts by jointly encoding each sub-question with candidate entities and their associated images. For answer inference, a hybrid reasoning module operates over the DMKG via two parallel paths: (1) relation-linking prediction; (2) RAG Reasoning with large vision-language models. A background-reflective decision module then aggregates evidence from both paths to select the most credible answer. Experimental results on MMQAKE show that Hybrid-DMKG significantly outperforms existing MKE approaches, achieving higher accuracy and improved robustness to knowledge updates.

\end{abstract}

%
\section{Introduction}

With the rapid advancement and widespread adoption of large language models (LLMs) \cite{zhao2023survey,achiam2023gpt,chang2024survey,10.1145/3726302.3729980,SHEN2025102860,zhang2025mars,zhang2025maps,wang2025model}, knowledge editing (KE) has emerged as a critical research area 
\begin{figure}    
  \centering
    \includegraphics[scale=.721]{ 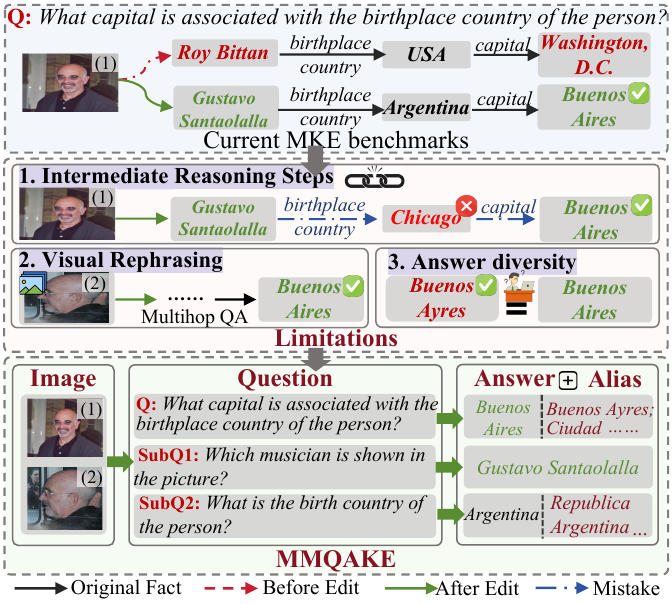}
  \caption{An example of our benchmark (MMQAKE), which differs in evaluation from existing MKE benchmarks.}
  \label{FIG:Introduction}
  \vspace{-6mm}
\end{figure}
\cite{meng2022locating,mengmass,mitchell2022memory}. KE aims to revise inaccurate, incomplete, or outdated knowledge encoded in LLMs while minimizing unintended alterations to unrelated content. To systematically evaluate whether such edits improve model responses, particularly for complex queries whose answers depend on the updated knowledge, \citet{zhong2023mquake} proposed the \emph{multihop question answering with knowledge editing} (MQUAKE) task.  MQUAKE requires models to perform multihop reasoning over modified knowledge and has been explored in text-only settings~\cite{gu2024pokemqa,shi2024retrieval,lu2025knowledge}. However, many real-world applications involve multimodal information, such as text, images, and videos, which present new challenges in fusing and representing diverse modalities~\cite{zhang2024mc, 10.5555/3737916.3738210}. This highlights the necessity of multimodal knowledge editing (MKE), an extension of KE that enables reasoning and modification both visual and textual modalities.

\begin{table*}
\small
\centering
\begin{tabular}{c c c c c c}
\toprule
\multirow{2}{*}{\textbf{Benchmark}} & 
\multirow{2}{*}{\textbf{Multimodal}} & 
\multirow{2}{*}{\textbf{Visual Rephrasing}} & 
\multicolumn{3}{c}{\textbf{Evaluation}} \\
\cmidrule(lr){4-6}
 & & & \textbf{Multihop Accuracy} & \textbf{Hop-wise Accuracy} & \textbf{Aliases} \\
\midrule
MQUAKE          & \xmark      &\xmark        & \cmark              & \cmark                           & \cmark   \\
VLKEB(multihop) & \cmark      & \xmark          & \cmark              &  \xmark                                 & \xmark   \\
MMQAKE         & \cmark      & \cmark          & \cmark              & \cmark                           & \cmark   \\
\bottomrule
\end{tabular}
\caption{The comparison of different benchmarks across various evaluation dimensions for multimodal multihop  QA. ``Visual Rephrasing'' refers to the use of alternative images of the same entity to evaluate multihop reasoning. ``Aliases'' refers to whether answer aliases are accepted as correct during evaluation. }

\label{different}
\vspace{-4mm}
\end{table*}

Despite recent progress in MKE \cite{10.5555/3737916.3738210,dummke}, current benchmarks primarily evaluate the correctness of final answers produced by large vision-language models (LVLMs) \cite{li2023blip, liu2023visual, zhu2024minigpt, DBLPCQ25,chen2025learning,liang2025seeing,10.1145/3746027.3755711}, while giving little attention to the quality of intermediate reasoning and robustness to visually rephrased inputs. For example, in Figure~\ref{FIG:Introduction}, although the person’s name is modified from ``\emph{Roy Bittan}" to ``\emph{Gustavo Santaolalla}", existing benchmarks still assess only the final answer “\emph{Buenos Aires}” for the multihop question $Q$, without examining the reasoning steps required to derive it. Such end-only evaluation risks masking reasoning errors~\cite{zhong2023mquake}, thereby limiting the reliability and interpretability of MKE performance. These issues manifest in three key limitations, as illustrated in Figure~\ref{FIG:Introduction}. (1) \textbf{Lack of accurate evaluation of intermediate reasoning steps.}
In multihop question answering, models may occasionally produce the correct final answer while relying on outdated or incorrect facts~\cite{gu2024pokemqa}, such as (\emph{Gustavo Santaolalla}, ``\textbf{birthplace country}", \emph{Chicago}). Ignoring the correctness of intermediate reasoning steps obscures the model’s actual reasoning process and undermines the reliability of the evaluation.
(2) \textbf{Lack of robustness evaluation under visual rephrasing.} Robust MKE methods should produce consistent outputs even when input images are visually modified (e.g., from image (1) to (2)). However, existing benchmarks often overlook this aspect, limiting the model’s ability to generalize to real-world scenarios where visual content may be modified or presented in diverse forms.
(3) \textbf{Neglect of valid alias diversity.} For example, answers such as \emph{Buenos Ayres} are not recognized as equivalent to \emph{Buenos Aires}, despite their semantic equivalence. This can penalize correct answers, undermining fair evaluation and potentially misrepresenting model performance.

To address these limitations, we propose Multimodal Multihop Question Answering with Knowledge Editing (MMQAKE), an extension of the VLKEB benchmark~\cite{10.5555/3737916.3738210}, as shown in Figure~\ref{FIG:Introduction}. MMQAKE features multihop questions requiring 2 to 5 reasoning steps, each aligned with a factual link in a reasoning chain. When multimodal knowledge is updated, models need to correctly propagate the revised information and generate answers that reflect the updated facts \cite{zhong2023mquake}. Besides, we evaluate predictions at each intermediate step \cite{gu2024pokemqa}, enabling fine-grained assessment of reasoning quality. Additionally, we include visually rephrased images to test robustness to visual variations. Finally, following the MQUAKE evaluation protocol, we consider all valid aliases of the ground-truth answer (e.g., \emph{Buenos Aires} and \emph{Buenos Ayres}), as retrieved from Wikidata.
The key differences between MMQAKE and existing benchmarks, including VLKEB and MQUAKE, are summarized in Table~\ref{different}.
Using MMQAKE, we further evaluate several representative MKE approaches to assess their effectiveness in complex reasoning scenarios. Our results reveal that many existing methods~\cite{chen-etal-2020-recall,zhu2020modifying,de2021editing,mitchell2022memory,zheng2023can}, struggle with the multihop and cross-modal challenges.

To address the faithfulness of current MKE methods in multihop question answering, we propose \textbf{Hybrid-DMKG}: a hybrid reasoning framework built upon dynamic multimodal knowledge graphs (DMKG). The DMKG represents knowledge as structured triples $(\emph{head}, \emph{relation}, \emph{tail})$, where entities are linked with corresponding images, and supports dynamic updates to accommodate evolving knowledge. This framework enriches semantic connections and enhances reasoning capabilities in LVLMs \cite{liang2025kag, li2024subgraphrag}.
Moreover, inspired by Chain-of-Thought reasoning~\cite{wei2022chain} and multihop question decomposition~\cite{zhong2023mquake, gu2024pokemqa}, we employ LLMs without fine-tuning to decompose multihop question into a sequence of sub-questions. 
For visual-based sub-questions, we utilize a multimodal retrieval model that jointly encodes the sub-question, candidate entities, and their associated images from the DMKG, with the goal of retrieving the entity most relevant to the sub-question as the answer. For reasoning-based sub-questions, we propose a hybrid reasoning module that operates along two parallel pathways to generate candidate answers: (1) relation-link prediction, which traverses the DMKG to infer an answer directly, and (2) retrieval-augmented generation–enhanced reasoning in the LVLM, which incorporates context from the DMKG. A background-reflective decision module then aggregates evidence from both paths to select the most credible answer. Our main contributions can be summarized as follows:

\begin{table*}[]
    \centering
    \small
    \begin{tabular}{cccccccc}
        \toprule
        \textbf{Datasets} & \textbf{Edit Number} & \textbf{2-hop} & \textbf{3-hop} & \textbf{4-hop} & \textbf{5-hop} & \textbf{Sub-question Number} & \textbf{Average Aliases} \\
        \midrule
        (MMQAKE) (Eval) & 1,278 & 1,278 & 1,238 & 1,193 & 1,110 & 11,773 & 9.49 \\
        \bottomrule
    \end{tabular}
\caption{Statistics of the MMQAKE dataset. The ``Average Aliases'' denotes the average number of answer aliases.} 
    \label{tab:mmquake_stats}
    \vspace{-5mm}
\end{table*}

\begin{itemize}
\item We propose \textbf{MMQAKE}, the first benchmark for multimodal multihop question answering with knowledge editing, extending the existing MKE tasks. MMQAKE challenges models to reason over both textual and visual modalities across 2 to 5-hop factual chains. In addition, it evaluates robustness to visual rephrasing in multihop questions, simulating real-world scenarios where knowledge must be accurately updated and reflected through complex reasoning.

\item We propose \textbf{Hybrid-DMKG}, a step-by-step reasoning framework built on a dynamic multimodal knowledge graph that continuously maintains and updates structured multimodal knowledge. By integrating complementary reasoning strategies, symbolic relation traversal, and retrieval-augmented generation in LVLM, this framework enhances the accuracy of multihop inference. Moreover, we propose a reflective decision module that effectively reconciles differing reasoning outputs, leading to more robust and faithful answers.

\item Extensive experiments on MMQAKE with multimodal knowledge editing methods reveal that most struggle with multihop and cross-modal reasoning. Our proposed Hybrid-DMKG approach significantly outperforms existing baselines, demonstrating higher accuracy and improved robustness to knowledge updates.

\end{itemize}





\section{Methodology}

\subsection{Problem Definition}


Multimodal knowledge editing is formalized as a quadruple $\mathcal{D} = ( x, v, o, \tilde{o})$, where $x$ is the textual input,  $v$ is corresponding a image, and the objective is to update a fact from $o$ to $\tilde{o}$. This editing operation is denoted as $f = ( x,v, o \to \tilde{o})$.  Based on this formulation, we introduce the task of \textbf{MMQAKE}, referred to as the textual MQUAKE task. 

Given a multihop question $Q$ associated with an image $v$, answering $Q$ requires executing a sequence of intermediate queries that form a multihop reasoning chain. This process can be represented as:
$
C = [\{o, r_{1}, y_{1}\}, \{t_{2}, r_{2}, y_{2}\}, \dots, \{t_{n}, r_{n}, y_{n}\}]
$. At the $k$-th hop, $t_{k}$ denotes the subject, $r_{k}$ the relation, and $y_{k}$ the object. Notably, the object of the $(k{-}1)$-th fact serves as the subject of the $k$-th fact, i.e., $y_{k-1} = t_{k}$.
In Figure~\ref{FIG:Introduction}, the initial set of relationships includes: ([IMAGE], \textit{name}, \textit{Roy Bittan
}), (\textit{Roy Bittan
}, \textit{birthplace country}, \textit{USA}), and (\textit{USA
}, \textit{capital}, \textit{Washington, D.C.}). Based on this chain, a 3-hop question such as \textit{What capital is associated with the birthplace country of the person?} can be formulated. When multimodal facts in the chain are edited ([IMAGE], \textit{name}, \textit{Roy Bittan
} $\rightarrow$ \textit{Gustavo 
Santaolalla}), LVLM leverages the updated knowledge to answer the multihop question correctly $y_n \rightarrow \tilde{y}_{n}$ (\textit{Washington, D.C.
} $\rightarrow$ \textit{Buenos Aires}).


Besides, we argue that an effective MKE method should incorporate all edits from the knowledge corpus $C$ into the model~\cite{gu2024pokemqa}, thereby enabling internal reasoning over the updated information. To evaluate whether these edits have been integrated, we assess the model’s ability to answer decomposed sub-questions derived from multihop queries, as illustrated in Figure~\ref{FIG:Introduction}. The model needs to correctly answer each sub-question to ensure consistency throughout the reasoning chain, i.e., $((y_1, y_2, \dots, y_{n-1}) \rightarrow (\tilde{y}_1, \tilde{y}_2, \dots, \tilde{y}_{n-1}))$. To further evaluate generalization in the visual modality, we test the model on both the original image $v$ and a visually rephrased image $\tilde{v}$ to evaluate the model’s robustness and generalization across related visual inputs under the edited knowledge setting.


\subsection{Dataset Construction}
\textbf{MMQAKE} extends VLKEB~\cite{10.5555/3737916.3738210} by evaluating models on \textbf{each step of multihop questions}, \textbf{visual rephrasing}, and \textbf{linguistic diversity in answers}, thereby increasing the complexity of reasoning depth and cross-modal understanding. Specifically, each multihop question in MMQAKE is augmented with three \emph{paraphrased questions} generated via the ChatGPT API to simulate natural language ambiguity \cite{gu2024pokemqa,lu2025knowledge}. Additionally, each question is \emph{decomposed into sub-questions}, each accompanied by paraphrases and annotated intermediate answers, enabling a \emph{step-by-step evaluation} of the model’s reasoning process. To evaluate robustness to visual rephrasing, we introduce alternative images from the VLKEB that depict the same entity as the original edited image. Finally, to ensure fair and semantically robust evaluation, we construct \emph{answer alias sets} based on Wikidata references, mitigating the impact of linguistic variation in answers. Dataset statistics are summarized in Table~\ref{tab:mmquake_stats}.

\begin{figure*}    
  \centering
    \includegraphics[scale=.61]{ 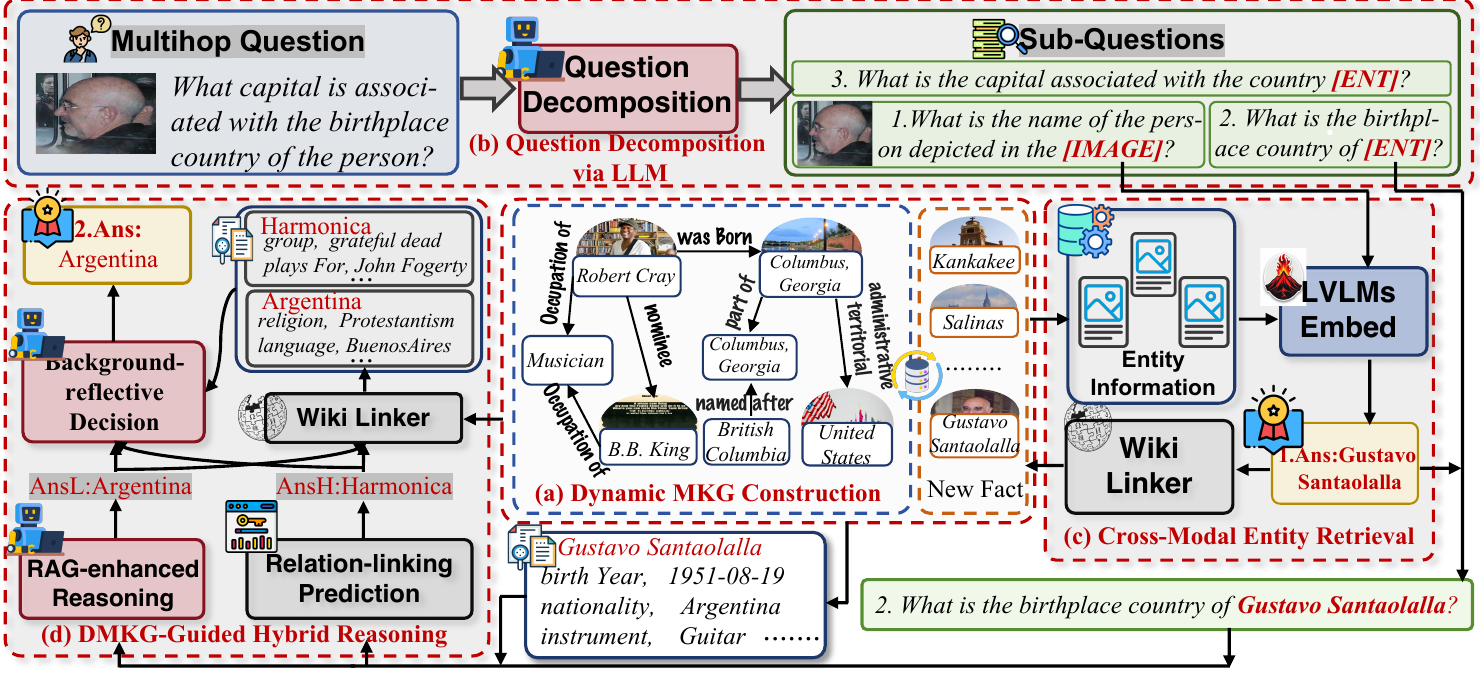}
  \caption{Overall framework of Hybrid-DMKG for MMQAKE task.}
  \label{FIG:Main}
  \vspace{-5mm}
\end{figure*}
\subsection{Hybrid-DMKG Framework}

Hybrid-DMKG is a parameter-preserving framework, which comprises the following key components: \textbf{(a) Dynamic MKG Construction:} We construct and maintain a structured DMKG, where knowledge is encoded as image-related triples. This structure enables efficient updates and deletions, supporting real-time adaptation to evolving facts. \textbf{(b) Question Decomposition:} We utilize an LLM to decompose multihop questions into multiple single-hop sub-questions, distinguishing between visual sub-questions that require image-based support and reasoning sub-questions that rely on structured knowledge. \textbf{(c) Cross-Modal Entity Retrieval from DMKG:} Visual sub-questions are handled using a cross-modal retrieval model that jointly encodes the visual query and each entity in the DMKG. The model then retrieves the most relevant entity as the answer. \textbf{(d) DMKG-Guided Hybrid Reasoning:} For reasoning sub-questions, candidate answers are generated through two parallel pathways: (1) relation linking within the DMKG to identify relevant answers, (2) an RAG-based method that enhances the LVLM’s generation using DMKG-derived context. Then, a reflective decision module jointly evaluates the supporting background knowledge retrieved from the DMKG for each candidate and selects the most plausible answer.

\subsubsection{Dynamic MKG Construction}



We use an MKG $\mathcal{G}$ as an external source to manage multimodal knowledge, providing a clear and editable structure for multihop knowledge updates and multihop traversal. As shown in Figure~\ref{FIG:Main}, each statement in $\mathcal{G}$ is represented as $({\mathcal{G}^e_{i}}, {\mathcal{G}^r_{i}}, {\mathcal{E}^{o}_i})$, where the head entity ${\mathcal{G}^e_{i}}$ and the tail entity ${\mathcal{G}^{o}_i}$ belong to the entity set $\mathcal{E}$. Some entity ${\mathcal{G}^e_{i}}$ is associated with a corresponding image ${\mathcal{G}^{v}_i}$. The relation ${\mathcal{G}^r_{i}}$ defines the semantic connection between the head and tail entities.

To incorporate new multimodal knowledge, we integrate an edit quadruple $(x,v, o, \tilde{o})$ into the MKG $\mathcal{G}$, resulting in the DMKG $\tilde{\mathcal{G}}$. The updated MKG $\tilde{\mathcal{G}}$ retains both the original and edited facts, enabling the model to reason over both prior and newly integrated multimodal information.

\subsubsection{Question Decomposition}

Inspired by prior work that decomposes multihop questions into sub-questions to improve retrieval accuracy~\cite{zhong2023mquake, gu2024pokemqa, lu2025knowledge,li-etal-2025-topology}, we employ an LLM without fine-tuning to decompose multimodal multihop questions. Specifically, we design a template $P_{\text{Dec}}$  that transforms a given multimodal multihop question  $Q$ into a set of sub-questions:
\begin{eqnarray}
\{q_{1}, q_{2}, \dots, q_{n}\} = \text{LLM}(Q, P_{\text{Dec}})
\label{eq:0}
\end{eqnarray}
As illustrated by the example in Figure~\ref{FIG:Main}, a multihop question is decomposed into three sub-questions. For sub-questions that involve visual information, the placeholder \texttt{[IMAGE]} is used to indicate a reference to the image. To promote entity consistency across sub-questions, related entities in other sub-questions are replaced with a special token \texttt{[ENT]}. After decomposing the main question, we sequentially address each sub-question.

\subsubsection{Cross-Modal Entity Retrieval from DMKG}
Unlike MQUAKE, which focuses exclusively on textual queries, MMQAKE also requires accurate identification of visual content.  For example, answering the sub-question $q_1$ requires the model to recognize the entity depicted in the rephrased image $\tilde{v}$. Inspired by multimodal retrieval methods~\cite{lewis2020retrieval,lin2024mm}, we employ a cross-modal retriever $\text{M}_{u}$ to perform retrieval across different modalities. Specifically, we treat the entities in the DMKG $\tilde{\mathcal{G}}$, each associated with an image and a name, as the candidate answer corpus. To enhance entity representation and retrieval accuracy, we integrate both the image and its linked entity name from $\tilde{\mathcal{G}}$ into a unified representation,
\begin{equation}\label{eq:1}
{z_m} = {\text{M}_{u}}([\tilde{\mathcal{G}}^e_m, \tilde{\mathcal{G}}^{v}_m])
\end{equation}
where $\tilde{\mathcal{G}}^e_m$ denotes the entity name of the $m$-th entity in $\tilde{\mathcal{G}}$ and $\tilde{\mathcal{G}}^v_m$ is its associated image. The sub-question $q_1$ and the rephrased input image $\tilde{v}$ are encoded using the same module:
\begin{equation}\label{eq:2}
{s} = {\text{M}_{u}}([{q}_1,\tilde{v}])
\end{equation} 
Then, we identify the most relevant entity $a_1$ as the subject of next sub-question by computing top1 similarity between query vector $s$ and candidate entity representations ${z^m}$:
\begin{equation} 
{a_1} = \mathop {\arg {\rm{Top1}}}\limits_{m \in \{ 1,2, \cdots M\} } \frac{{{{({s})}^T}{z_m}}}{{{{\left\| {{s}} \right\|}_2}{{\left\| {{z_m}} \right\|}_2}}}
\label{eq:3}
\end{equation} 
where $M$ denotes the total number of entities with corresponding images in the DMKG. As shown in Figure~\ref{FIG:Main}, the retrieved answer $a_1$ is \emph{Gustavo Santaolalla}. We replace the \texttt{[ENT]} in $q_2$ with $a_1$ to form the next-step sub-question: \emph{What is the country of birth of Gustavo Santaolalla?}.

{
\renewcommand{\arraystretch}{0.85}
\begin{table*}[ht]
\centering
\small
\setlength{\tabcolsep}{1.6mm}
\begin{tabular}{
>{\centering\arraybackslash}m{2.2cm}|
>{\centering\arraybackslash}m{2.5cm}|
>{\centering\arraybackslash}m{1.1cm}|
>{\centering\arraybackslash}m{1.6cm}|
>{\centering\arraybackslash}m{1.2cm}|
>{\centering\arraybackslash}m{1.2cm}|
>{\centering\arraybackslash}m{1.2cm}|
>{\centering\arraybackslash}m{1.2cm}|
>{\centering\arraybackslash}m{1.2cm}
}
\toprule
\textbf{Input Image} & \textbf{Backbones} & \textbf{Metrics} & \textbf{FT(QFor)} & \textbf{FT(All)} & \textbf{MEND} & \textbf{SERAC} & \textbf{IKE} & \textbf{Ours} \\
\midrule
\multirow{6}{*}{\makecell[c]{Original \\ Image}} 
 & \multirow{2}{*}{\makecell[c]{BLIP-2 (3.8B)}}     & M-Acc & 3.73      &   0.32      &  0.04   &    5.75   &  16.64    &   \textbf{47.55}   \\
 &                            & H-Acc & 0.20        &   0.02      &   0.00  &  0.00     &   6.16   &  \textbf{28.88}    \\
\cline{2-9}
 & \multirow{2}{*}{\makecell[c]{LLaVA  (7B)}}     & M-Acc & 4.63      &    1.66     &  0.70   &    6.58   &   38.93   &   \textbf{53.75}   \\
 &                            & H-Acc & 0.44     &   0.00      &   0.00  &    0.00   &   16.38   &    \textbf{29.90}  \\
\cline{2-9}
 & \multirow{2}{*}{\makecell[c]{MiniGPT-4 (7.8B)}}   & M-Acc & 4.69         &    0.08     &  0.07   &    0.27   &   15.48   &   \textbf{35.86}   \\
 &                            & H-Acc & 0.44       &    0.00     &  0.00   &   0.00    &  6.14    & \textbf{24.73}     \\
\cline{1-9}
\multirow{6}{*}{\makecell[c]{Rephrased \\ Image}} 
 & \multirow{2}{*}{\makecell[c]{BLIP-2 (3.8B)}}     & M-Acc & 0.84       &   0.04      &  0.02   &   1.04    &  14.44    &   \textbf{45.27}   \\
 &                            & H-Acc & 0.11    &   0.00    &  0.00 &  0.00   &  6.06    &     \textbf{26.08}  \\
\cline{2-9}
 & \multirow{2}{*}{\makecell[c]{LLaVA (7B)}}     & M-Acc & 5.71     &    1.61     &  0.06   &  0.97     &   37.61  &   \textbf{51.27}   \\
 &                            & H-Acc & 0.53     &   0.00      &   0.00  &  0.02     &   16.91   &   \textbf{26.16}  \\
\cline{2-9}
 & \multirow{2}{*}{\makecell[c]{MiniGPT-4 (7.8B)}}  & M-Acc & 4.17       &     0.11    &  0.04   &    0.13   &   9.86   &    \textbf{33.41}  \\
 &                            & H-Acc & 0.77    &     0.00    &   0.00  &   0.00    &   5.76   &  \textbf{22.23}    \\
\bottomrule
\end{tabular}
\caption{Experimental results (\%) on the MMQAKE dataset. “QFor” and “All” refer to fine-tuning only the Q-Former parameters and all model parameters, respectively. The best results are highlighted in \textbf{bold}.}
\label{tab:ft-comparison}
\vspace{-4mm}
\end{table*}
}

\subsubsection{DMKG-Guided Hybrid Reasoning}

In reasoning sub-questions, such as $q_2$, we further utilize the DMKG to improve the accuracy and interpretability of answer generation by the LVLM. To retrieve related knowledge from $\tilde{\mathcal{G}}$, we first address variability in natural language expressions (e.g., \emph{United States of America} vs. \emph{USA}, need to refer to the same entity). We apply a \textbf{Wiki Linker} \footnote{\url{https://wiki.osdev.org/Linker}} module $\phi$ to normalize the entity ${a_{1}}$, mapping it to its canonical form $e_{2}$ within $\tilde{\mathcal{G}}$,
\begin{equation}\label{eq:5}
e_{2} = \phi(a_{1})
\end{equation}
Based on the linked entity, we extract its associated triples from the $\tilde{\mathcal{G}}$ as a knowledge set,
\begin{equation}\label{eq:5:2}
C_2 = \varphi(e_{2},\tilde{\mathcal{G}})
\end{equation}
where $\varphi$ denotes the operation that retrieves all relational triples from the DMKG associated with the given entity $e_2$. The resulting set of associated knowledge is defined as $C_2 = \left\{ (e_2, \tilde{\mathcal{G}}^r_{e_2,j}, \tilde{\mathcal{G}}^o_{e_2,j}) \;\middle|\; j = 1, \ldots, k \right\}$. As illustrated in Figure~\ref{FIG:Main}, (\emph{birth year}, \emph{1951-08-19}) for the entity \emph{Gustavo Santaolalla}.
After obtaining the related knowledge set $\mathcal{C}_2$, we propose a hybrid reasoning module with three parts: (1) \emph{Relation-Link Prediction}, (2) \emph{RAG-Enhanced Reasoning in LVLM}, which together generate candidate answers, and (3) a \emph{Background-Reflective Decision} module that aggregates these candidates to select the most credible response.

\noindent
\textbf{(1) Relation-linking Prediction} This module leverages explicit DMKG's relational information for answer prediction. It performs graph-based reasoning over relational paths by assessing the semantic similarity between the query and candidate relation types. Based on our observation, many queries can be answered directly by identifying the underlying relational intent expressed in the question. For example, in query $q_2$ about \textbf{Gustavo Santaolalla}, the implicit relation keyword is ``\emph{country of birth}". If a semantically related relation exists in the DMKG, the most relevant entity can be retrieved and is highly likely to serve as the answer. Motivated by this, we introduce a fine-tuned relation extractor $\text{M}_{e}$ to identify explicit relational keywords $k^q_2$ from the query $q_2$:
\begin{equation}\label{key_word}
k^q_2 = \text{M}_{e}(q_2)
\end{equation}

The extracted relational keyword $k^q_2$ is then encoded in an embedding $h(k^q_2)$ using a lightweight word embedding \emph{Sense2Vec} \cite{trask2015sense2vec}. Given the candidate answer set $C_2$ extracted from the knowledge graph, we compute the cosine similarity between the query keyword embedding $h(k^q_2)$ and each candidate relation embedding $h(\tilde{\mathcal{G}}^r_{e_2,j})$. The candidate answer with the highest similarity score is selected as follows:
\begin{equation}\label{eq:4}
\begin{aligned}
j^* &= \arg\max_j \cos\big(h(k^q_2), h(\tilde{\mathcal{G}}^r_{e_2,j})\big) \\
a^{\text{link}}_{2} &=
\begin{cases}
\tilde{\mathcal{G}}^o_{e_2,j*}, & \text{if } \cos\big(h(k^q_2), h(\tilde{\mathcal{G}}^r_{e_2,j*})\big) \geq \alpha \\
\varnothing, & \text{otherwise}
\end{cases}
\end{aligned}
\end{equation}
where $j^*$ denotes the index of the candidate relation that is most semantically aligned with the query. If the similarity score exceeds a threshold $\alpha$, the corresponding object $\tilde{\mathcal{G}}^o_{e_2,j*}$ is selected as the predicted answer $a^\text{link}_2$. Otherwise, if no relevant relation is identified, the answer is indicated by $\varnothing$.

\noindent
\textbf{(2) RAG-enhanced Reasoning in LVLM}
While the linking prediction module is generally effective in identifying target entities and their associated relations, this method may fail when background knowledge is incomplete or when key term extraction is inaccurate. To address this limitation, inspired by the use of retrieval-augmented generation in LLMs~\cite{he2024g, shi2024retrieval} for enhanced reasoning, we propose a RAG-enhanced reasoning module based on DMKG. Specifically, we retrieve the top$K$ knowledge snippets $\mathcal{K}_{\text{Ret}}$ from the associated triple set $\mathcal{C}_2$ that are semantically most relevant to the current query $q_2$. These retrieved snippets are then incorporated into the answer prompt $P_\text{Ans}$ and provided as input to the LVLM. This design allows the LVLM to access external knowledge, thereby enhancing its reasoning capabilities when faced with incomplete or ambiguous information,
\begin{equation}\label{eq:6}
\small
a^{\text{model}}_2 = \text{LVLM}\left(q_2, \tilde{v}, \mathcal{K}_{\text{Ret}}\left(q_2,C_2\right),\ P_{\text{Ans}}\right)
\end{equation}
where $a^{\text{model}}_2$ denotes the output of LVLM with RAG. $\mathcal{K}_{\text{Ret}}$ uses the same model architecture as described in Equations~(\ref{eq:1})–(\ref{eq:3}), with the key distinction that only the textual modality is employed.

\noindent
\textbf{(3) Background-reflective Decision}
In certain cases, the candidate answers produced by the two reasoning paths differ, i.e., $a_2^{\text{link}} \neq a_2^{\text{model}}$. To resolve such conflicts, we propose a background-reflective decision module. Instead of relying solely on initial predictions, this module enables LVLM to reflectively evaluate competing answers by leveraging the rich semantic and relational context provided by the DMKG. Specifically, for each candidate answer, we extract background information based on the adjacency of the entity level in the DMKG, as determined by the entity link mechanism defined in Equations~(\ref{eq:5})–(\ref{eq:5:2}). The contextual background knowledge representations $C^\text{link}_2$ and $C^\text{modal}_2$, corresponding to $a^{\text{link}}_2$ and $a^{\text{modal}}_2$, are defined as follows:
\begin{equation}\label{eq:9}
\begin{aligned}
\big\{ C^\text{link*}_2,\; C^\text{modal*}_2 \big\} = 
\varphi \Big( \phi \big( \{ a^{\text{link}}_2,\; a^{\text{modal}}_2 \} \big) , \tilde{\mathcal{G}}\Big)
\end{aligned}
\end{equation}
These contexts encompass relevant entity descriptions and co-occurring facts that collectively enhance the model's reasoning capabilities. The final prediction is generated by the LVLM using the original question $q_2$, the associated input image $\tilde{v}$, and answer-related background knowledge as input. This process is formalized as:
\begin{equation}
\small
a_2 = \text{LVLM}(q_2, \tilde{v}, [a^\text{link}_2, C^\text{link*}_2], [a^\text{modal}_2, C^\text{modal*}_2], P_{\text{Cho}})
\end{equation}
where $a_2$ denotes the final answer after reflective decision, and $P_{\text{Cho}}$ is the choice prompting strategy guiding the reflective reasoning process. In multihop reasoning, we repeat Equations (2)–(4) for sub-questions requiring cross-modal entity retrieval, and apply Equations (5)–(8) during answer generation. By dynamically invoking relevant modules at each reasoning hop, the model gradually resolves multihop questions and generates a final answer.

\section{Experiments}
We conducted comparative experiments to evaluate the performance of the proposed Hybrid-DMKG method against several existing approaches. Additional details (e.g., \textbf{evaluation metrics}, \textbf{experimental setup}, \textbf{more experimental results}, \textbf{case studies}, and \textbf{prompt templates}) are provided in Appendix~A–H. The implementation is publicly available at https://github.com/YuanLi95/Hybrid-DMKG.

\subsection{Results}
\subsubsection{Overall Results}
Table~\ref{tab:ft-comparison} presents our experimental results on the MMQAKE dataset. With the exception of IKE, most existing MKE methods exhibit significant performance degradation on multihop question answering tasks. Notably, MEND performs the worst, failing to complete any multihop reasoning task, despite demonstrating strong performance on standard single-hop editing tasks~\cite{10.5555/3737916.3738210}. Moreover, increasing model size does not lead to improved performance. For instance, MiniGPT-4 frequently underperforms compared to the smaller BLIP-2 model. In contrast, IKE, a retrieval-augmented method, maintains relatively stable baseline performance. However, it struggles to integrate multihop information effectively, leading to a significant decline in H-Acc as the number of editing rounds increases.

Our proposed Hybrid-DMKG framework consistently outperforms all baseline methods across various evaluation metrics and backbone configurations. When employing BLIP-2 as the backbone LVLM, Hybrid-DMKG achieves an H-Acc score that surpasses IKE by 22.72\% on original images, highlighting its better capability in addressing MMQAKE. Furthermore, with LLaVA as the backbone, Hybrid-DMKG attains M-Acc and H-Acc scores of 53.75\% and 29.90\%, respectively, demonstrating strong generalizability across different architectures. Notably, the rephrased-image setting introduces significant challenges for multimodal generalization. Under this condition, most models exhibit varying degrees of performance degradation, IKE with  MiniGPT-4 showing particularly pronounced declines. While Hybrid-DMKG also faces increased difficulty in cross-modal retrieval due to the introduction of rephrased images, it consistently outperforms other benchmark models. This robustness is primarily attributed to the incorporation of a dynamic multimodal knowledge graph, which enriches contextual understanding. Additionally, our proposed hybrid reasoning framework enables parallel reasoning along two distinct pathways and integrates their insights through a reflective decision-making module, resulting in more accurate and reliable outputs.

\begin{figure}    
  \centering
  \includegraphics[scale=.366]{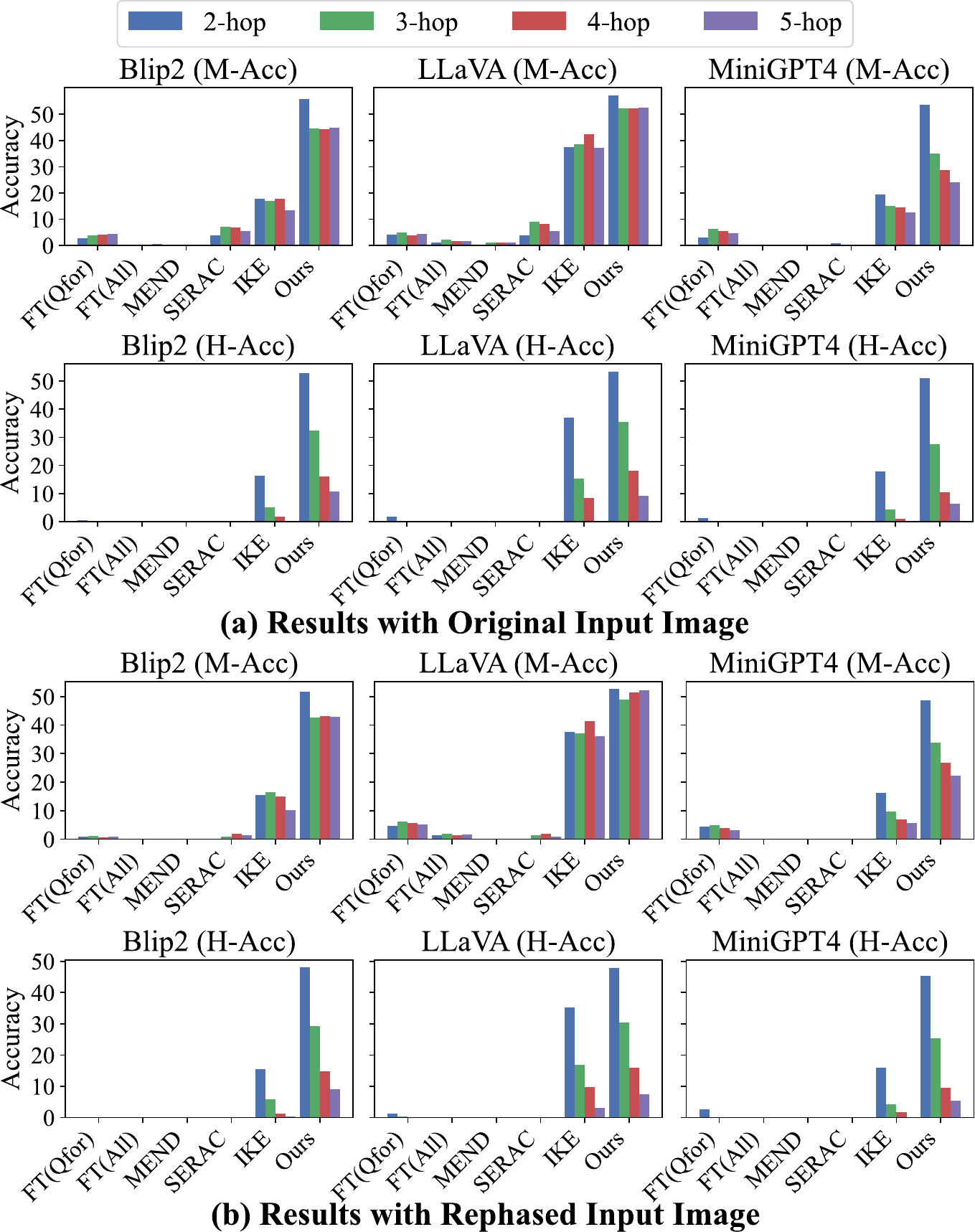}
  \caption{Performance comparison of different hops on MMQAKE using the original and rephrased input images.}
  \label{FIG:multi-result}
  \vspace{-5mm}
\end{figure}

\subsubsection{Results for Different Hops}

We further analyzed model performance across varying hop counts (2 to 5). As shown in Figure~\ref{FIG:multi-result}, under the M-Acc metric, all models maintain relatively stable performance regardless of hop count. This suggests that multiple valid reasoning paths can lead to correct final answers. However, this stability may also reflect a limitation of the M-Acc metric itself, which only evaluates the final answer correctness and may fail to capture differences in reasoning quality or path validity across varying hop lengths. In contrast, the H-Acc metric imposes a stricter requirement: every intermediate reasoning step must be correct. Consequently, performance consistently declines as the hop count increases. Notably, our model significantly outperforms baselines on 4-hop and 5-hop questions under H-Acc, achieving nearly double their accuracy. In the most challenging 5-hop setting, Hybrid-DMKG exceeds 5\% accuracy, while other methods typically remain below 2\%. This improvement stems from our effective multihop question decomposition, which mitigates error propagation and reduces hallucinations by large language models. Additionally, the cross-modal retrieval component boosts step-wise accuracy by supplying relevant candidate answers, thereby strengthening the overall reasoning process.

\begin{table}[]
\setlength{\tabcolsep}{.15mm}
\small
\begin{tabular}{clcccccc}
\toprule
\multirow{2}{*}{\textbf{\begin{tabular}[c]{@{}c@{}}Input \\ Image\end{tabular}}} & \multirow{2}{*}{\textbf{Models}} & \multicolumn{2}{c}{\textbf{BLIP-2}} & 
\multicolumn{2}{c}{\textbf{LLaVA}} &
\multicolumn{2}{c}{\textbf{MiniGPT-4}}  \\
 &                         & M-Acc       & H-Acc       & M-Acc         & H-Acc        & M-Acc       & H-Acc       \\
 \midrule
\multirow{4}{*}{{{\begin{tabular}[c]{@{}c@{}}Original \\ Image\end{tabular}}}} &Ours                    &  47.55           &     28.88        &    53.75             &      29.90      &   35.86          &   24.73          \\
 & w/o \emph{Linking}  &     46.09        &   18.59          &    47.68           &   23.15           &  24.13           &     14.13           \\
& w/o \emph{RAG} 
& \multicolumn{6}{>{\columncolor[rgb]{0.906, 0.855, 0.824}}c}{28.13 \enspace   21.50} \\

& w/o \emph{Decision} &     48.19        &        28.05     &     52.71          &    28.36          &      30.44        &    20.23         \\
\midrule
\multirow{4}{*}{{{\begin{tabular}[c]{@{}c@{}}Rephrased \\ Image\end{tabular}}}}   &Ours                     &  45.27           &    26.08         & 51.27              &    26.16          &   33.41          &      22.23       \\
 & w/o \emph{Linking}  &  37.22          &       15.85      &       43.13       &   17.95            &  21.08          &     8.30   \\
& w/o \emph{RAG} 
& \multicolumn{6}{>{\columncolor[rgb]{0.906, 0.855, 0.824}}c}{26.13 \enspace19.41}
\\
 & w/o \emph{Decision} &      44.18       &       23.76      &    46.75           &   23.37           &  28.53         &   16.58          \\ 
\bottomrule
\end{tabular}
\caption{Ablation study results. The \emph{w/o RAG} setting denotes that only the linking prediction module is used to obtain the answer. As a result, all LVLM backbones yield identical performance under this configuration.}
\label{Ablation study results}
\vspace{-4mm}
\end{table}

\subsection{Ablation Study}


To assess the contribution of each component in the Hybrid-DMKG framework, we conducted ablation studies on three core modules: Relation-linking Prediction (\emph{Linking}), RAG-enhanced Reasoning in LVLM (\emph{RAG}), and Background-reflective Decision (\emph{Decision}). As shown in Table~\ref{Ablation study results}, removing the \emph{Linking} module from MiniGPT-4 leads to a substantially larger performance drop than removing the \emph{RAG} module. For instance, under the rephrased input image setting, the H-Acc score declined by 13.93\%. This suggests that MiniGPT-4 struggles to effectively incorporate information from the DMKG, often generating semantically incoherent or irrelevant responses. In contrast, LLaVA exhibits stronger knowledge aggregation and reasoning capabilities, enabling the \emph{RAG} module to function more effectively and achieve better overall performance. Although the removal of the \emph{Decision} module does not completely disable the model, it results in a significant performance degradation, particularly when processing rephrased input images. This highlights the importance of the background-reflective decision module in filtering out incorrect candidate answers by leveraging relevant background knowledge. Accordingly, the \emph{Decision} module enhances the robustness of decision-making and improves the accuracy of the final responses.

\section{Related Work}

\subsection{Multimodal Knowledge Editing Methods}
Current MKE methods typically adapt existing LLM editing techniques \cite{touvron2023llama} by modifying specific neural network layers and fall into two main categories. (1) \textbf{Parameter-update methods} integrate new knowledge into model parameters, such as fine-tuning and MEND \cite{de2021editing}, which approximates gradient updates via low-rank decomposition. While effective, these approaches risk catastrophic forgetting, incur high training costs, and can degrade model performance, especially in multihop reasoning \cite{gu-etal-2024-model}. (2) \textbf{Parameter-retention methods} preserve model parameters and influence outputs through external mechanisms like in-context learning. For example, SERAC \cite{mitchell2022memory} uses a scope classifier and counterfactual modeling, while IKE \cite{zheng2023can} employs demonstrations to guide edits. However, these methods often depend on task-specific, single-hop textual supervision, limiting their generalization to multihop or cross-modal reasoning.
In contrast to existing approaches, Hybrid-DMKG leverages MKG to enable dynamic knowledge updates and retrieval without requiring modification of model parameters. By integrating cross-modal retrieval, Hybrid-DMKG effectively addresses multimodal reasoning tasks that involve both textual and visual inputs. Moreover, we propose a hybrid reasoning module that generates answers from parallel reasoning paths, combined with reflective decision-making mechanisms, which further enhance the accuracy and reliability of the responses.

\subsection{Multihop QA with Knowledge Editing}

Recently, to more comprehensively evaluate the reasoning capabilities of KE methods, \citet{zhong2023mquake} introduced MQUAKE. Unlike traditional KE benchmarks, which primarily assess updates by verifying edited facts or answering single-hop factual queries, MQUAKE emphasizes the model’s ability to perform multihop reasoning after knowledge has been injected or updated. This evaluation paradigm is better aligned with the complex reasoning demands typical of real-world \cite{yuan2023joint,gu2024pokemqa,shi2024retrieval,lu2025knowledge}. Recent approaches that integrate RAG with question decomposition have demonstrated strong performance in both knowledge editing and reasoning tasks, offering promising directions for advancing the field \cite{gu2024pokemqa,shi2024retrieval,lu2025knowledge,li2025meta,ijcai2025p908}.

However, these methods are not directly applicable to MMQAKE, which requires cross-modal knowledge editing and reasoning. To address this, Hybrid-DMKG builds on prior work \cite{shi2024retrieval,sun2024alice,sun2024assessing} with two key enhancements: (1) a cross-modal retrieval model that jointly encodes text and images for accurate entity recognition and multimodal knowledge localization; and (2) a hybrid reasoning module that integrates relation-linking prediction with RAG-enhanced LVLM generation to produce complementary answers. A background-reflective decision module then evaluates these answers using external knowledge, enhancing response consistency and reliability.

\section{Conclusion}
In this paper, we introduce MMQAKE, the first benchmark of multimodal multihop question answering with knowledge editing, expanding existing multimodal knowledge editing benchmarks. MMQAKE features questions requiring 2-5 reasoning steps in both textual and visual modalities, and an evaluation protocol that checks the factual consistency in every reasoning stage. To address this task, we propose Hybrid-DMKG, a hybrid reasoning framework built upon a dynamic multimodal knowledge graph that enables continual knowledge updates. Hybrid-DMKG combines traditional relation-based prediction with RAG using LVLMs to produce parallel answers. A reflective-decision module is used to enhance cross-modal inference and harmonize divergent reasoning outcomes. Extensive experiments demonstrate that our approach significantly outperforms existing methods on the MMQAKE benchmark.

In future work, we plan to extend MMQAKE to support dynamic knowledge updates by incorporating temporal and event-based information. Additionally, we aim to address open-ended questions beyond factoid QA and explore end-to-end multihop reasoning without relying on predefined sub-questions.

\section{Acknowledgments}
This research is supported by the National Natural Science Foundation of China (62476097, 62276072), the Fundamental Research Funds for the Central Universities, South China University of Technology (x2rjD2250190),  Guangdong Provincial Fund for Basic and Applied Basic Research—Regional Joint Fund Project (Key Project) (2023B1515120078), Guangdong Provincial Natural Science Foundation for Outstanding Youth Team Project (2024B1515040010), Guangdong Basic and Applied Basic Research Foundation (2025A1515010162), Guangxi Natural Science Foundation Key Project (No. 2025GXNSFDA069017). TW was supported by the NIHR Maudsley Biomedical Research Centre, Maudsley Charity, King’s Together, MHaPS Early Career Researcher Award, and MRC DATAMIND.

\bibliography{aaai2026}
\section{Appendix}

\subsection{Appendix A: Evaluation Metrics}
To comprehensively evaluate the effectiveness of knowledge editing models on multihop question answering involving edited knowledge, we adopt the evaluation metrics proposed in MQUAKE: Multihop Accuracy (\textbf{M-Acc}) \cite{zhong2023mquake} and Hop-wise Accuracy (\textbf{H-Acc}) \cite{gu2024pokemqa}. For both \textbf{M-Acc} and \textbf{H-Acc}, a generated answer is considered correct only if it exactly matches one of the elements in the reference gold answer set.

\textbf{M-Acc} evaluates only the correctness of the final answer. However, this metric may overestimate a model’s reasoning capability, as the model might arrive at the correct answer through an incorrect reasoning path. It thus obscures the model’s actual reasoning and fails to assess reasoning faithfulness. In contrast, \textbf{H-Acc} assesses the correctness of each intermediate answer in the reasoning chain,  mitigating the risk of false positives caused by reasoning shortcuts. Thus, \textbf{H-Acc} serves as the primary evaluation metric in our study, as it more accurately reflects the model’s ability to apply edited knowledge throughout the entire reasoning process. Importantly, an instance is considered incorrect under \textbf{H-Acc} if any single step in the reasoning path is incorrect.

\subsection{Appendix B: Backbone Models and Baselines}
We evaluate current representative MKE methods using three widely adopted LVLM backbones: BLIP-2 (3.8B)~\cite{li2023blip}, LLaVA-1.5 (7B) ~\cite{liu2023visual}, and MiniGPT-4 (7.8B)~\cite{zhu2024minigpt}.

\subsubsection{Baselines} 
Following prior work on MKE~\cite{zheng2023can,10.5555/3737916.3738210}, we adopt several baselines that fall into two categories: \textbf{parameter-update} methods and \textbf{parameter-retention} methods.  

\textbf{Parameter-update methods} modify the model’s internal parameters. \textbf{Fine-tune} \cite{chen-etal-2020-recall,zhu2020modifying} involves updating the components of the model, including both the LLM layers and the vision module. \textbf{MEND} \cite{de2021editing} updates the final layers of the LLM within LVLM by applying low-rank gradient decomposition combined with predictive parameter updates. In contrast, \textbf{parameter-retention methods} preserve the original model parameters. \textbf{SERAC} \cite{mitchell2022memory} is a memory-based approach composed of a classifier and a counterfactual model. In our implementation, the classifier is based on BERT, while the counterfactual model is adapted to each LVLM by aligning it with the corresponding LLM architecture. \textbf{IKE}~\cite{zheng2023can} retrieves semantically similar examples from the training data to construct and inject new knowledge, with this retrieval-based editing strategy applied uniformly across all models.




\begin{table}
\centering
\setlength{\tabcolsep}{.8mm}
\begin{tabular}{ccccc}
\toprule
\multirow{2}{*}{Hyperparameters} & epochs& learning rate & batch size & optimizer\\
& 10        & 2e-5 &128      & AdamW \\
\bottomrule
\end{tabular}
\caption{Hyperparameters for training the extractor $M_e$.}
\label{tab:parameter}
\end{table}

\begin{table}
\centering
\setlength{\tabcolsep}{.8mm}
\begin{tabular}{cccc}
\toprule
\multirow{2}{*}{Datasets Statistics} & Total Number & Question Lengths & Labels \\
& 10216        & 8.80             & 1.43 \\
\bottomrule
\end{tabular}
\caption{Statistics of the dataset used for relational keyword. The dataset is split into training, development, and test sets in a 6:3:1 ratio.}
\label{tab:relation-data-statistics}
\vspace{-4mm}
\end{table}
\subsection{Appendix C: Experimental Setup}


\begin{table*}
\centering
\small
\setlength{\tabcolsep}{1mm}
\begin{tabular}{cccccccccccc}
\toprule
\multirow{2}{*}{\textbf{Input Image}} & \multirow{2}{*}{\textbf{Backbones}} 
& \multicolumn{2}{c}{\textbf{2-hop}} & \multicolumn{2}{c}{\textbf{3-hop}} 
& \multicolumn{2}{c}{\textbf{4-hop}} & \multicolumn{2}{c}{\textbf{5-hop}} 
& \multicolumn{2}{c}{\textbf{All}} \\
& & M-Acc & H-Acc & M-Acc & H-Acc & M-Acc & H-Acc & M-Acc & H-Acc & M-Acc & H-Acc \\
\midrule
\multirow{3}{*}{\begin{tabular}[c]{@{}c@{}}Original\\ Image\end{tabular}} 
& BLIP-2       & 23.01 & 21.18 & 16.84 & 11.29 & 14.72 & 11.29 & 14.53 & 2.61  & 17.17 & 9.68 \\
& LLaVA        & 27.99 & 25.38 & 22.45 & 14.37 & 22.17 & 8.79  & 19.70 & 4.03  & 23.01 & 12.99 \\
& MiniGPT-4    & 24.43 & 23.49 & 16.03 & 11.18 & 16.31 & 3.53  & 19.75 & 2.43  & 18.92 & 9.91 \\
\midrule
\multirow{3}{*}{\begin{tabular}[c]{@{}c@{}}Rephrased\\ Image\end{tabular}} 
& BLIP-2       & 22.83 & 21.45 & 17.28 & 11.38 & 15.31 & 4.61  & 14.44 & 2.62  & 17.36 & 9.81 \\
& LLaVA        & 23.02 & 21.23 & 20.38 & 11.82 & 20.01 & 4.85  & 18.41 & 2.02  & 20.45 & 9.98 \\
& MiniGPT-4    & 21.29 & 20.05 & 15.29 & 10.02 & 16.55 & 3.23  & 18.60 & 2.35  & 17.83 & 8.81 \\
\bottomrule
\end{tabular}
\caption{Experimental results (\%) of Hybrid-DMKG under the no-alias evaluation setting.}
\label{tab:no_alis_hop-performance}
\end{table*}
\begin{table*}[]
\small
\centering
\setlength{\tabcolsep}{1.0mm}
\begin{tabular}{lcccc ccc ccc c}
\toprule
\multicolumn{2}{c}{\textbf{Modules}} & \multicolumn{2}{c}{\textbf{BLIP-2}} & \multicolumn{2}{c}{\textbf{LLaVA}} & 
\multicolumn{2}{c}{\textbf{MiniGPT-4}} & \multirow{2}{*}{\textbf{I-Acc}} \\
\cmidrule(lr){3-4} \cmidrule(lr){5-6} \cmidrule(lr){7-8}
Retrieval Model &  Decomposition Model & M-Acc & H-Acc & M-Acc & H-Acc & M-Acc & H-Acc & \\
\midrule
CLIP (428M) & Gemini &\textbf{47.55}& \textbf{28.88} & 53.75 &29.90 & \textbf{35.86} & \textbf{24.73} & \textbf{63.87} \\
\midrule
CLIP (428M) & LLaMA2-7B   &43.72 & 23.39  &47.47 &23.19 &31.59 &19.98 &62.17  \\
 CLIP (428M)   & ChatGPT     & \underline{46.06} & \underline{27.08} & 51.91 & 27.79 &\underline{34.71} &\underline{22.78}   &\underline{63.50} \\
MM-Embed (7B) & Gemini&45.44&26.12 & \textbf{55.44} & \textbf{30.67} &  33.45 &22.45 & 58.37\\

MM-Embed (7B) & LLaMA2-7B   &42.35 & 21.08  &48.88 &24.09 & 29.11&17.14 &53.97  \\
MM-Embed (7B)   & ChatGPT     & 45.19 & 24.88 & \underline{54.36} & \underline{30.23} &32.16 &20.46 &57.31 \\

\bottomrule
\end{tabular}
\caption{Comparison of question decomposition and cross-modal retrieval under both LLM and LVLM settings using the original image. I-Acc denotes first-hop image retrieval accuracy. \textbf{Bold} and \underline{underlined} indicate the best and second-best results, respectively. ``428M’’ and ``8B’’ refer to the parameter sizes of CLIP and MM-Embed, respectively.}
\label{tab:model_comparison}
\vspace{-4mm}
\end{table*}

We trained a lightweight relation extraction model based on DistilBERT~\cite{sanh2019distilbert} using the MQUAKE dataset~\cite{zhong2023mquake}, which is derived from Wikidata\footnote{\url{https://www.wikidata.org/wiki/Wikidata}}. Detailed training configurations and data construction procedures are provided in Table~\ref{tab:parameter} and Table~\ref{tab:relation-data-statistics}.
To address entity disambiguation, we employ Wiki Linker to align textual mentions with their corresponding Wikipedia entries. In the relation-linking prediction module, we set the similarity score threshold $\alpha$ to 0.5. For RAG-enhanced reasoning in LVLMs and background-reflective decision-making, the number of retrieved knowledge entries is set to 5.

For question decomposition, we utilized both open- and closed-source LLMs without fine-tuning, including LLaMA2-7B~\cite{touvron2023llama}, GPT-3.5-turbo-0125~\cite{achiam2023gpt}, and Gemini 2.5 Flash-Thinking\footnote{\url{https://deepmind.google/models/gemini/}}. In the retrieval module, we adopt CLIP~\cite{radford2021learning}, a lightweight multimodal pretraining model, as the retrieval framework to support diverse cross-modal retrieval tasks between text and images. For the initial multimodal knowledge graph, we used MKG~\cite{liu2019mmkg} as the base resource. Following data cleaning and knowledge updates, the resulting graph contains 58,542 entities, of which 11,087 have corresponding images, yielding a total of 686,048 triples. All experiments were conducted on a server equipped with 5$\times$NVIDIA L40-48G GPUs. For training parameter-preserving methods (e.g., MEND and IKE),  we use the original VLKEB training set, consisting of 5,000 knowledge editing examples for training the classifiers.

\subsection{Appendix D: Performance under the no-alias evaluation}


By incorporating an alias set into our evaluation protocol, we mitigate bias caused by linguistic variation, enabling a more comprehensive and fair assessment of model performance and avoiding potential underestimation. To illustrate the importance of alias-aware evaluation, we report in Table~\ref{tab:no_alis_hop-performance} the performance of our proposed method with the alias set removed. The results show that excluding alias handling leads to a significant drop in performance, underscoring the necessity of accounting for expression diversity.

For instance, when evaluating LLaVA using both original and rephrased image descriptions, its H-Acc drops significantly, from 29.90\% and 26.16\% to 12.99\% and 9.98\%, respectively. Notably, BLIP-2 performs worse than MiniGPT-4 on the original inputs, and the H-Acc gap between LLaVA and other models narrows considerably. These performance shifts highlight the limitations of conventional VQA evaluation protocols, which rely on exact single-answer matching and thus fail to capture nuanced differences in model capabilities. Incorporating an alias set not only enhances evaluation robustness but also provides a more accurate and comprehensive reflection of model performance.

\subsection{Appendix E: Comparative Study of Alternative Modules}

Since our method leverages LLMs without fine-tuning for question decomposition and employs a cross-modal model to retrieve image and background knowledge from the MKG, we conducted systematic evaluations of these two components independently. The results using original images as input are presented in Table~\ref{tab:model_comparison}, while additional results using rephrased images are provided in Table~\ref {tab:model_rephrased}.

In the question decomposition stage, we evaluated LLMs of varying scales and architectures, including LLaMA2-7B, GPT-3.5-turbo-0125 (ChatGPT), and Gemini 2.5 Flash-Thinking (Gemini). Experimental results indicate that LLaMA2-7B significantly degrades overall system performance, highlighting its limited capacity to construct complex reasoning chains. In particular, inadequate logical decomposition substantially hinders the effectiveness of multimodal retrieval, resulting in a 1.70\% decline in first-hop cross-modal retrieval accuracy and subsequently reducing the quality of the generated answers. In contrast, both ChatGPT and Gemini demonstrate stronger logical reasoning capabilities. Notably, Gemini achieves the highest H-Acc score, largely due to its pretraining on datasets rich in logical reasoning and diverse inference patterns, which enhances its generalization ability in complex decomposition and reasoning tasks.
\begin{table*}[ht!]
\small
\centering
\setlength{\tabcolsep}{1.0mm}
\begin{tabular}{lcccc ccc ccc c}
\toprule
\multicolumn{2}{c}{\textbf{Modules}} & \multicolumn{2}{c}{\textbf{BLIP-2}} & \multicolumn{2}{c}{\textbf{LLaVA}} & 
\multicolumn{2}{c}{\textbf{MiniGPT-4}} & \multirow{2}{*}{I-Acc} \\
\cmidrule(lr){3-4} \cmidrule(lr){5-6} \cmidrule(lr){7-8}
Retrieval Model &  Decomposition Model & M-Acc & H-Acc & M-Acc & H-Acc & M-Acc & H-Acc & \\
\midrule
CLIP (428M) & Gemini &\textbf{45.27}& \textbf{26.08} & \textbf{51.27} & \textbf{26.16} & \textbf{33.41} & \textbf{22.22} & \textbf{55.01} \\
\midrule
CLIP (428M) & LLaMA2-7B   &41.41 & 21.24  &44.28 &20.29 &30.15 &18.30 &52.93  \\
CLIP (428M)   & ChatGPT     & 43.29 & 24.90 & 49.51 & 25.17 &33.13 &21.48   &53.97 \\
\bottomrule
\end{tabular}
\caption{Comparison of question decomposition across LLMs using the rephrased image.}
\label{tab:model_rephrased}
\end{table*}
\begin{figure*}    
  \centering
    \includegraphics[scale=.67]{ 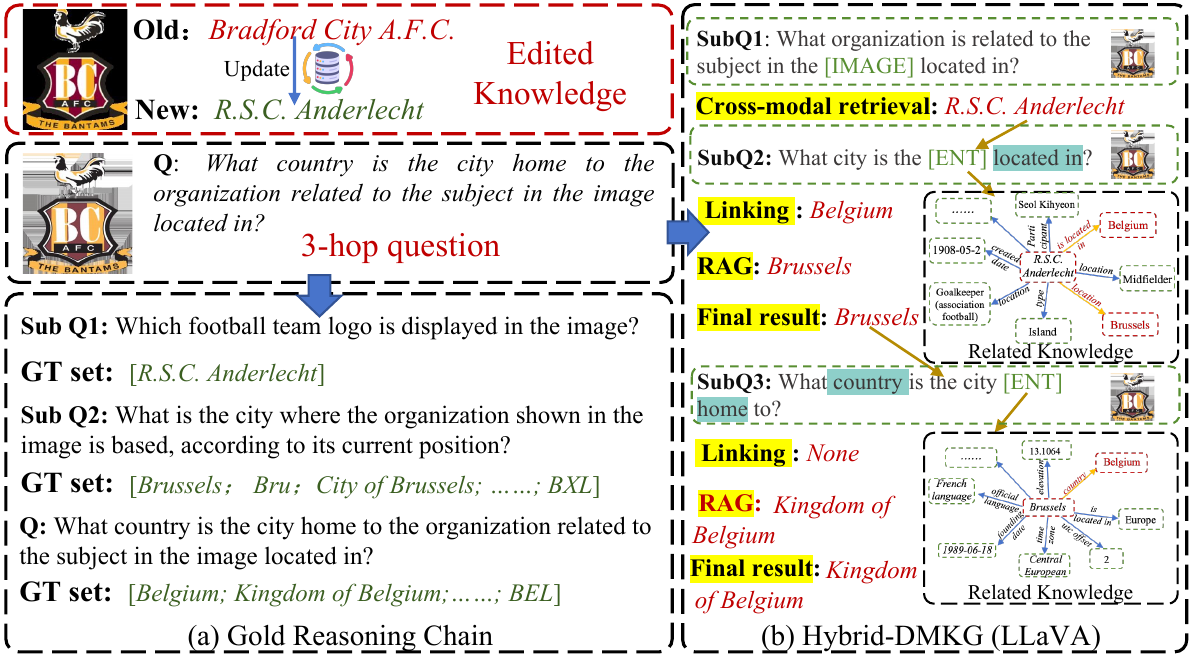}
\caption{A case study of Hybrid-DMKG solving a 3-hop question from MMQAKE. The phrase highlighted in blue (e.g., “located in”) represents the extracted relation keyword. “Linking” and “RAG” refer to the outputs of the Relation-Linking Prediction and RAG-Enhanced Reasoning modules within the LVLM, respectively. The ``GT set” denotes the ground truth set.}
  \label{FIG:case_study}
  \vspace{-4mm}
\end{figure*}

When evaluating the cross-modal retrieval module, we examined the impact of replacing the original CLIP with the larger MM-Embed (8B) model \cite{lin2024mm}, the state-of-the-art general-purpose model for cross-modal retrieval. The swap produced little improvement in first-hop cross-modal retrieval. However, using MM-Embed within LLaVA leads to markedly better results. This suggests that MM-Embed's enhanced text encoding capabilities significantly improve retrieval effectiveness during the subsequent text-only knowledge retrieval phase. Considering MM-Embed’s substantial parameter count, we default to the lighter CLIP model to maintain a better performance–efficiency trade-off.

Besides, Table~\ref{tab:model_rephrased} reports the performance of various decomposition models on rephrased images. We observe that visual rephrasing reduces the H-Acc performance gap across different models. Nevertheless, Gemini continues to achieve the highest overall performance, due to its better sub-question generation quality.

\subsection{Appendix F: Case Study}

We conducted a case study to evaluate the effectiveness of our method on complex multimodal question answering. Figure~\ref{FIG:case_study}(a) presents a 3-hop question from the MMQAKE set, along with its decomposed sub-questions that form a coherent reasoning chain. Figure~\ref{FIG:case_study}(b) outlines the solution process: Hybrid-DMKG first updates the MKG with new knowledge, and then an LLM decomposes the complex question into solvable sub-questions.

In the first sub-question, the framework performs cross-modal entity alignment. Thanks to its robust retrieval capabilities, it accurately identifies the target entity, even with rephrased visual input. The second sub-question requires reasoning over geographic information. The Linking Prediction module extracts the relation ``\emph{located in}" but incorrectly selects \textbf{Belgium}. In contrast, the RAG module, augmented with background knowledge, generates the candidate \textbf{Brussels}, which is correctly selected by the background-reflective decision module.

In the final sub-question, the Linking Prediction module correctly identifies the key term home but fails to map it to any relation in the DMKG, resulting in no retrieved answer (\emph{None}). In contrast, the RAG module successfully infers the correct answer, \emph{Kingdom of Belgium}, using both the knowledge retrieved and its own prior information. This answer aligns with the ground truth set constructed using aliases and is therefore considered correct.
These results demonstrate that Hybrid-DMKG effectively performs entity recognition and multihop reasoning across sub-questions, enabling accurate resolution of complex multimodal queries.

\subsection{Appendix G: Original Data for Different Hops Across Models}
Table~\ref{tab:detail_hop} presents the detailed results corresponding to Figure 3, covering different hop configurations on MMQAKE with both the original and rephrased input images.

\subsection{Appendix H: Details of Multihop Results for Ablation Study}

In the ablation study, we summarize the overall results, with detailed hop-wise performance reported in Table~\ref{table:multihop_abl}. The results demonstrate that removing \emph{RAG in LVLM} leads to a significantly larger performance drop as the number of hops increases. This suggests that as tasks require more external knowledge, relying solely on single-step \emph{linking prediction} becomes insufficient to retrieve the necessary information from the DMKG, thereby impairing the model’s ability to generate accurate answers. Additionally, we observe that MiniGPT-4 consistently yields the poorest performance in multihop scenarios, with the performance gap widening as the number of hops increases.

\subsection{Appendix I: Details on the Prompt Templates}
\begin{figure*}    
  \centering
  \includegraphics[scale=0.70]{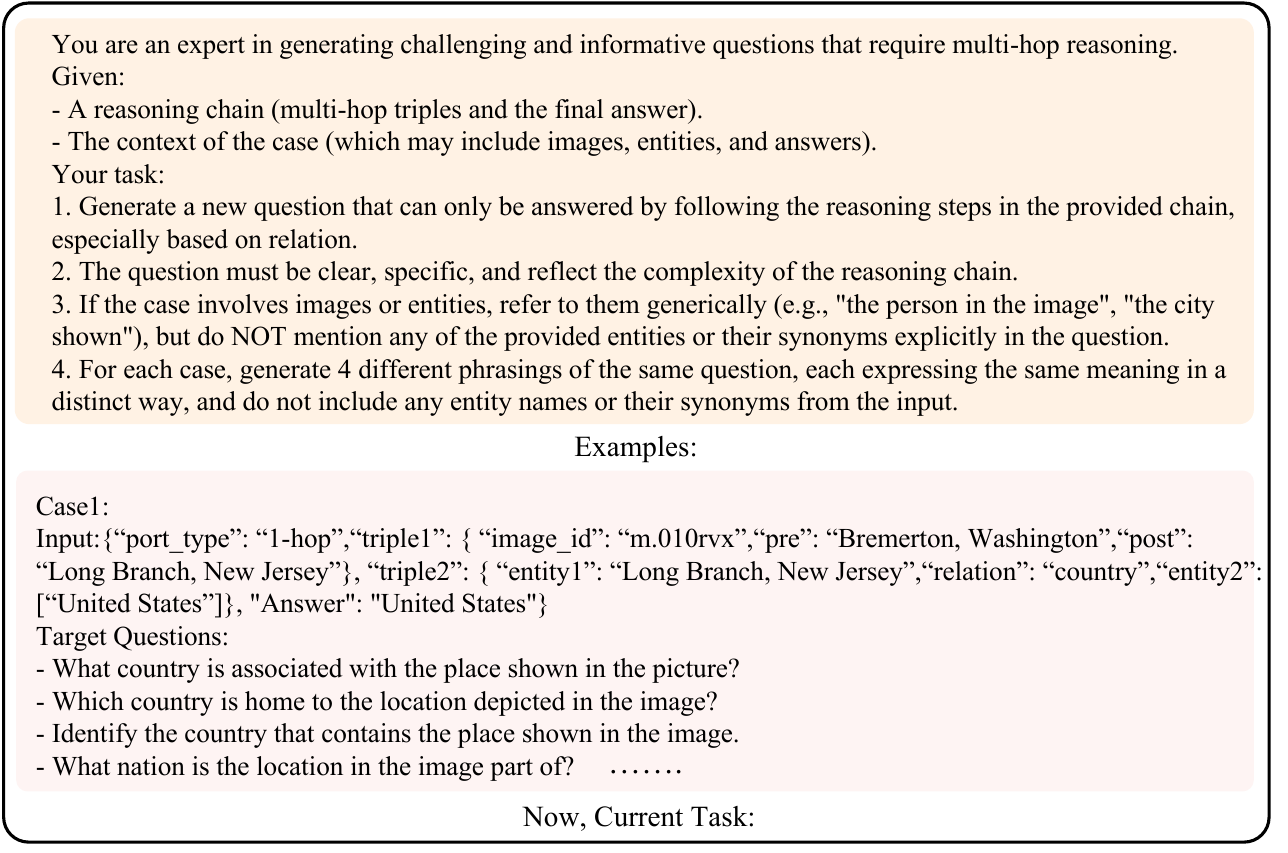}
\caption{Prompt template used for generating multihop questions in the MMQAKE task datasets.}
  \label{FIG:generate_task}
\end{figure*}

Figure~\ref{FIG:generate_task} shows the prompt template used by ChatGPT to generate the MMQAKE task datasets. For each question, ChatGPT produces four paraphrased versions, from which the first three are selected.

\begin{figure*}    
  \centering
  \includegraphics[scale=0.70]{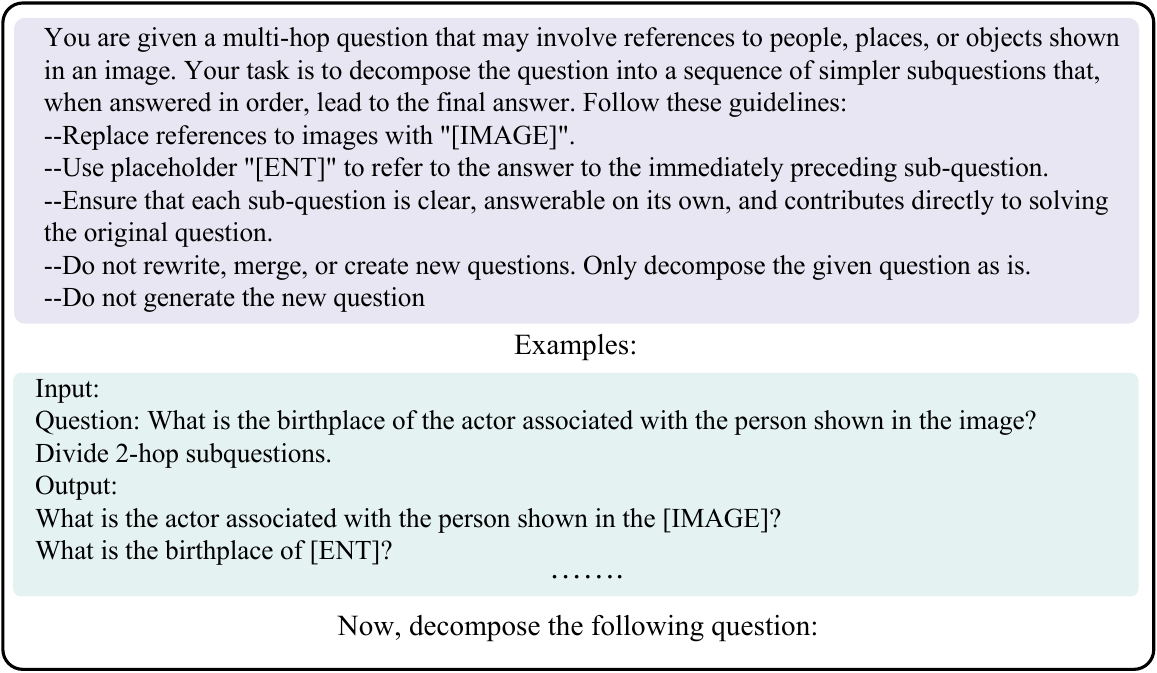}
\caption{Prompt template for decomposing multihop questions $P_\text{Dec}$.}
  \label{FIG:div}
\end{figure*}

\begin{figure*}    
  \centering
  \includegraphics[scale=0.70]{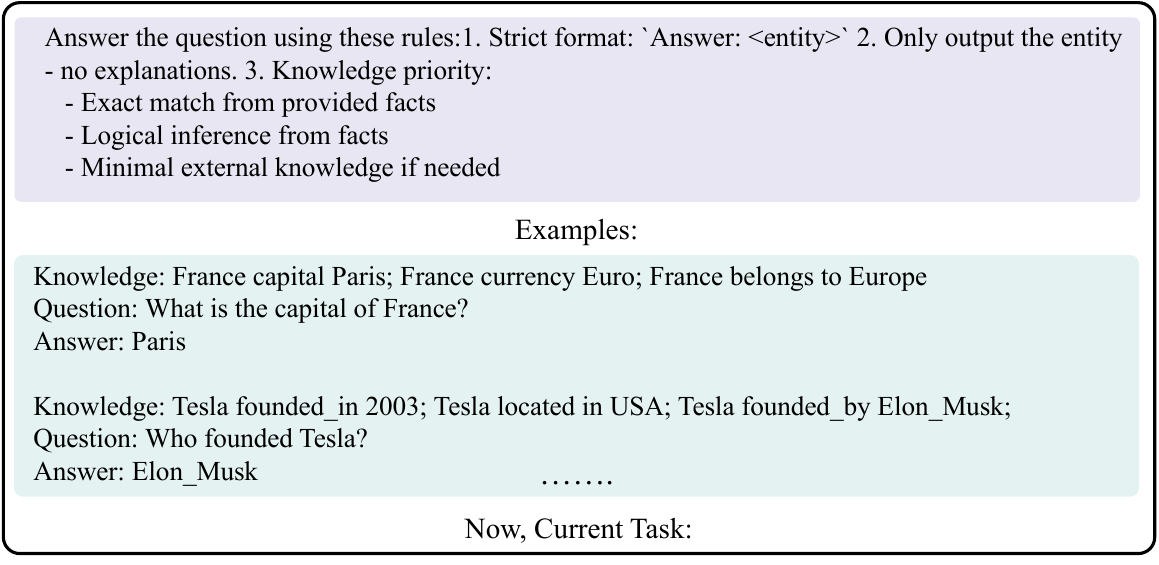}
\caption{Prompt template for the answer prompt $P_\text{Ans}$ used in RAG-enhanced reasoning.}
  \label{FIG:answer}
\end{figure*}

\begin{figure*}    
  \centering
  \includegraphics[scale=0.70]{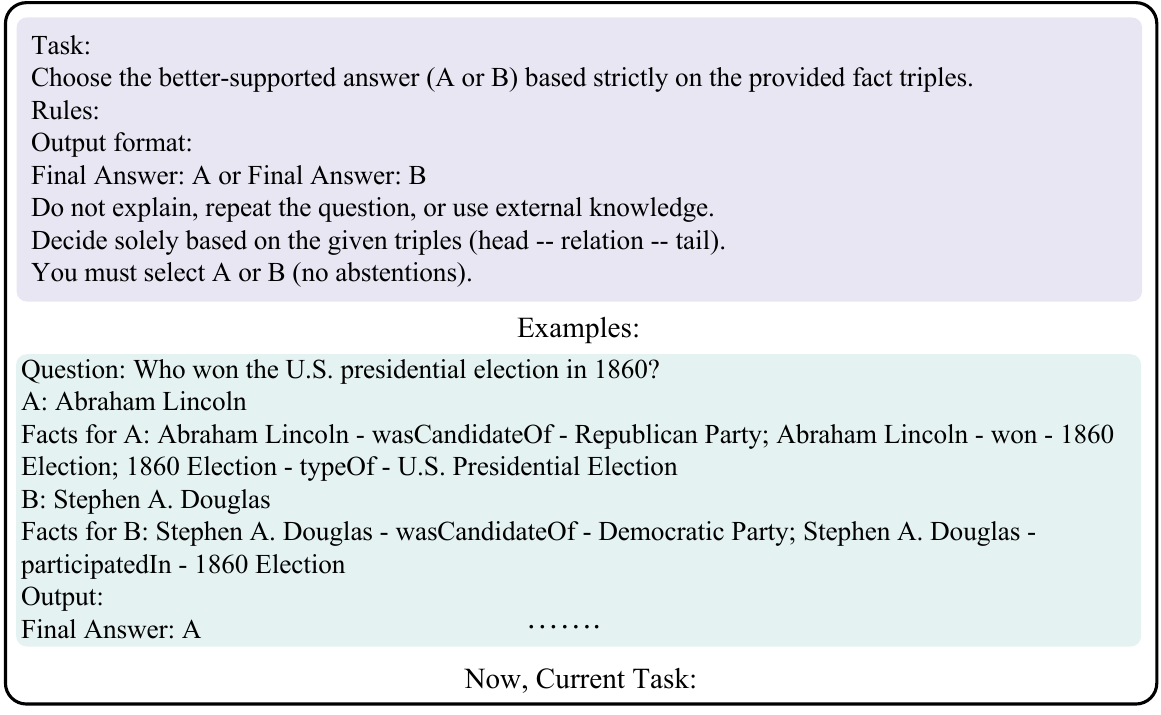}
  \caption{Prompt template for selecting the final answer $P_\text{Cho}$ from candidate answers.}
  \label{FIG:choice}
\end{figure*}

Figure~\ref{FIG:div} presents the prompt used to generate sub-questions from the original multihop questions. Figure~\ref{FIG:div} illustrates the prompt template employed to perform multihop question decomposition. Figures~\ref{FIG:answer} and~\ref{FIG:choice} illustrate the prompt templates used by LVLMs to (i) answer each decomposed sub-question and (ii) select the final answer from a set of candidates by leveraging background knowledge for reflective decision-making.

\begin{table*}[]
\centering
\small
\begin{tabular}{lllcccccccc}
\toprule
\multirow{2}{*}{\textbf{Input Image}} & 
\multirow{2}{*}{\textbf{Backbones}} & 
\multirow{2}{*}{\textbf{Models}} & 
\multicolumn{2}{c}{\textbf{2-hop}} & 
\multicolumn{2}{c}{\textbf{3-hop}} & 
\multicolumn{2}{c}{\textbf{4-hop}} & 
\multicolumn{2}{c}{\textbf{5-hop}} \\
& & & M-Acc & H-Acc & M-Acc & H-Acc & M-Acc & H-Acc & M-Acc & H-Acc \\
\midrule

\multirow{18}{*}{{{\begin{tabular}[c]{@{}c@{}}Original \\ Image\end{tabular}}}}
& \multirow{6}{*}{BLIP-2}     
& FT(Qformer) & 2.74 & 0.55 & 3.96 & 0.00 & 4.10 & 0.00 & 4.41 & 0.00 \\
& & FT(all)     & 0.31 & 0.08 & 0.33 & 0.00 & 0.25 & 0.00 & 0.45 & 0.00 \\
& & MEND        & 0.16 & 0.00 & 0.00 & 0.00 & 0.00 & 0.00 & 0.00 & 0.00 \\
& & SERAC       & 3.76 & 0.00 & 7.11 & 0.00 & 6.79 & 0.00 & 5.40 & 0.00 \\
& & IKE         & 17.92 & 16.43 & 17.04 & 5.09 & 17.77 & 1.84 & 13.51 & 0.18 \\
& & Ours        & 55.79 & 52.81 & 44.58 & 32.39 & 44.26 & 16.09 & 44.86 & 10.81 \\
\cline{2-11}

& \multirow{6}{*}{LLaVA}     
& FT(Qformer) & 4.23 & 1.72 & 4.85 & 0.24 & 3.77 & 0.00 & 4.41 & 0.00 \\
& & FT(all)     & 1.17 & 0.00 & 2.10 & 0.00 & 1.76 & 0.00 & 1.53 & 0.00 \\
& & MEND        & 0.23 & 0.00 & 1.05 & 0.00 & 1.17 & 0.00 & 0.99 & 0.00 \\
& & SERAC       & 3.76 & 0.00 & 9.05 & 0.00 & 8.13 & 0.00 & 5.41 & 0.00 \\
& & IKE         & 37.48 & 37.01 & 38.53 & 15.27 & 42.41 & 8.55 & 37.30 & 2.25 \\
& & Ours        & 57.27 & 53.44 & 52.26 & 35.54 & 52.25 & 18.02 & 52.61 & 9.28 \\
\cline{2-11}

& \multirow{6}{*}{MiniGPT4}  
& FT(Qformer) & 2.90 & 1.17 & 6.22 & 0.16 & 5.53 & 0.00 & 4.69 & 0.00 \\
& & FT(all)     & 0.16 & 0.00 & 0.00 & 0.00 & 0.00 & 0.00 & 0.18 & 0.00 \\
& & MEND        & 0.16 & 0.00 & 0.00 & 0.00 & 0.08 & 0.00 & 0.00 & 0.00 \\
& & SERAC       & 0.78 & 0.00 & 0.24 & 0.00 & 0.00 & 0.00 & 0.00 & 0.00 \\
& & IKE         & 19.32 & 17.84 & 14.94 & 4.28 & 14.50 & 1.01 & 12.71 & 0.27 \\
& & Ours        & 53.60 & 51.09 & 35.05 & 27.54 & 28.67 & 10.56 & 24.05 & 6.49 \\
\midrule

\multirow{18}{*}{{{\begin{tabular}[c]{@{}c@{}}Rephrased\\Image\end{tabular}}}}
& \multirow{6}{*}{BLIP-2}     
& FT(Qformer) & 0.86 & 0.23 & 1.05 & 0.00 & 0.58 & 0.00 & 0.91 & 0.00 \\
& & FT(all)     & 0.08 & 0.00 & 0.00 & 0.00 & 0.00 & 0.00 & 0.09 & 0.00 \\
& & MEND        & 0.00 & 0.00 & 0.00 & 0.00 & 0.08 & 0.00 & 0.00 & 0.00 \\
& & SERAC       & 0.16 & 0.00 & 0.81 & 0.00 & 1.92 & 0.00 & 1.35 & 0.00 \\
& & IKE         & 15.57 & 15.42 & 16.47 & 5.98 & 15.00 & 1.34 & 10.27 & 0.27 \\
& & Ours        & 51.59 & 48.04 & 42.56 & 29.40 & 43.18 & 14.90 & 42.97 & 9.09 \\
\cline{2-11}

& \multirow{6}{*}{LLaVA}     
& FT(Qformer) & 4.69 & 1.17 & 6.14 & 0.32 & 5.70 & 0.00 & 5.14 & 0.00 \\
& & FT(all)     & 1.48 & 0.00 & 1.86 & 0.00 & 1.51 & 0.00 & 1.53 & 0.00 \\
& & MEND        & 0.08 & 0.00 & 0.08 & 0.00 & 0.08 & 0.00 & 0.00 & 0.00 \\
& & SERAC       & 0.16 & 0.08 & 1.37 & 0.00 & 1.82 & 0.00 & 0.82 & 0.00 \\
& & IKE         & 37.61 & 35.21 & 37.24 & 16.96 & 41.41 & 9.72 & 36.21 & 3.06 \\
& & Ours        & 52.66 & 47.81 & 48.86 & 30.45 & 51.38 & 15.92 & 52.25 & 7.47 \\
\cline{2-11}

& \multirow{6}{*}{MiniGPT4}  
& FT(Qformer) & 0.46 & 2.66 & 4.92 & 0.24 & 4.02 & 0.00 & 3.15 & 0.00 \\
& & FT(all)     & 0.16 & 0.00 & 0.08 & 0.00 & 0.00 & 0.00 & 0.18 & 0.00 \\
& & MEND        & 0.16 & 0.00 & 0.00 & 0.00 & 0.00 & 0.00 & 0.00 & 0.00 \\
& & SERAC       & 0.23 & 0.00 & 0.00 & 0.00 & 0.17 & 0.00 & 0.18 & 0.00 \\
& & IKE         & 16.20 & 16.04 & 9.69 & 4.28 & 7.05 & 1.73 & 5.77 & 0.18 \\
& & Ours        & 48.60 & 45.46 & 33.92 & 25.44 & 26.90 & 9.55 & 22.34 & 5.49 \\
\bottomrule
\end{tabular}
\caption{The original data of different hop configurations on MMQAKE using both original and rephrased input images.}
\label{tab:detail_hop}
\end{table*}

\begin{table*}[]
\centering
\small
\setlength{\tabcolsep}{1.5mm}
\begin{tabular}{ccccccccccc}

\toprule
\multirow{2}{*}{\textbf{Input Image}} & \multirow{2}{*}{\textbf{Backbones}} & \multirow{2}{*}{\textbf{Models}} & \multicolumn{2}{c}{\textbf{2-hop}} & \multicolumn{2}{c}{\textbf{3-hop}} & \multicolumn{2}{c}{\textbf{4-hop}} & \multicolumn{2}{c}{\textbf{5-hop}} \\

&     &   & M-Acc & H-Acc & M-Acc & H-Acc & M-Acc & H-Acc & M-Acc & H-Acc \\
\midrule
\multirow{7}{*}{\begin{tabular}[c]{@{}c@{}}Original\\ Image\end{tabular}} 
& \multirow{2}{*}{BLIP-2}     & w/o \textit{Linking Prediction}  & 42.73 & 31.53 & 44.02 & 22.85 & 48.62 & 12.07 & 49.55 & 5.94  \\
&                            & w/o \textit{Reflective Decision} & 56.25 & 51.95 & 47.25 & 30.13 & 49.04 & 16.60 & 46.04 & 10.54 \\
\cline{2-11}
& \multirow{2}{*}{LLaVA}     & w/o \textit{Linking Prediction}  & 46.63 & 43.04 & 45.80 & 26.17 & 49.46 & 14.00 & 49.10 & 6.76  \\
&                            & w/o 
\textit{Reflective Decision} & 57.67 & 52.43 & 50.66 & 32.88 & 49.96 & 16.85 & 52.34 & 8.02  \\
\cline{2-11}
& \multirow{2}{*}{MiniGPT-4} & w/o \textit{Linking Prediction}  & 42.65 & 36.93 & 23.02 & 13.33 & 16.35 & 3.02  & 12.43 & 0.72  \\
&                            & w/o \textit{Reflective Decision} & 49.76 & 44.76 & 29.97 & 19.47 & 21.29 & 8.97  & 18.56 & 4.95  \\
\cline{2-11}
& \multicolumn{2}{c}{w/o \textit{RAG in LVLM}\rule{-1pt}{2.2ex}}               & 50.54 & 52.97 & 30.37 & 24.23 & 16.09 & 4.36  & 10.00 & 2.43  \\
\midrule
\multirow{7}{*}{\begin{tabular}[c]{@{}c@{}}Rephrased\\ Image\end{tabular}} 
& \multirow{2}{*}{BLIP-2}     & w/o \textit{Linking Prediction}  & 32.63 & 28.64 & 36.27 & 18.01 & 40.15 & 9.22  & 40.45 & 5.85  \\
&                            & w/o \textit{Reflective Decision} & 49.61 & 44.37 & 19.12 & 28.19 & 45.60 & 14.17 & 45.68 & 8.47  \\
\cline{2-11}
& \multirow{2}{*}{LLaVA}     & w/o \textit{Linking Prediction}  & 39.35 & 34.67 & 39.58 & 19.71 & 46.02 & 10.27 & 48.29 & 5.05  \\
&                            & w/o \textit{Reflective Decision} & 46.95 & 44.37 & 43.78 & 26.17 & 45.10 & 12.41 & 41.62 & 7.84  \\
\cline{2-11}
& \multirow{2}{*}{MiniGPT-4} & w/o \textit{Linking Prediction}  & 39.75 & 16.35 & 18.26 & 8.72  & 13.41 & 4.61  & 10.99 & 2.52  \\
&                            & w/o \textit{Reflective Decision} & 45.31 & 40.61 & 28.84 & 14.38 & 21.88 & 5.03  & 16.04 & 3.78  \\
\cline{2-11}
& \multicolumn{2}{c}{w/o \textit{RAG in LVLM}\rule{-1pt}{2.2ex}} & 48.90 & 45.77 & 27.79 & 21.41 & 15.42 & 5.20  & 9.64  & 2.16  \\

\bottomrule
\end{tabular}
\caption{Ablation results across 2–5-hop reasoning.}
\label{table:multihop_abl}
\end{table*}

\end{document}